\RequirePackage[svgnames]{xcolor}

\documentclass[11pt,logo,letterpaper]{berkeley}

\usepackage[utf8]{inputenc} %
\usepackage[T1]{fontenc}    %
\usepackage{lmodern}
\usepackage{textcomp}
\usepackage[svgnames]{xcolor}
\usepackage{asymptote}
\usepackage[comma,authoryear,compress]{natbib}
\usepackage{hyperref}
\usepackage{algorithm}
\usepackage{algorithmicx}
\usepackage{algpseudocode}
\usepackage{microtype}
\usepackage{graphicx}
\expandafter\def\csname ver@subfig.sty\endcsname{}
\usepackage{booktabs} %
\usepackage{float}
\usepackage{bigstrut}
\usepackage{tikz}
\usetikzlibrary{tikzmark}
\makeatletter
\newcommand*\myfontsize{%
  \@setfontsize\myfontsize{7}{8}%
}
\makeatother
\definecolor{myred}{rgb}{0.7, 0.3, 0.0}
\definecolor{myblue}{HTML}{054488}
\definecolor{mygreen}{HTML}{056b34}

\newcolumntype{R}[1]{>{\raggedleft\let\newline\\\arraybackslash\hspace{0pt}}m{#1}}

\usetikzlibrary{intersections}

\definecolor{darkgreen}{rgb}{0.0, 0.42, 0.24}

\usepackage{amsthm}
\usepackage{nicefrac}
\usepackage{subcaption}
\usepackage{caption}
\usepackage{graphicx}
\usepackage{cleveref}
\usepackage{bxcoloremoji}
\usepackage{listings}
\usepackage{float}
\usepackage{longtable}
\lstdefinestyle{python}{
    language=Python,
    basicstyle=\ttfamily\footnotesize,
    keywordstyle=\color{blue}\bfseries,
    commentstyle=\color{green},
    stringstyle=\color{red},
    numberstyle=\tiny\color{gray},
    showstringspaces=false,
    frame=single,
    breaklines=true,
    backgroundcolor=\color{lightgray!20}
}

\usepackage{hyperref}       %
\usepackage{url}            %
\usepackage{booktabs}       %
\usepackage{nicefrac}       %
\usepackage{microtype}      %
\usepackage{graphicx}
\usepackage{subcaption} 
\usepackage{wrapfig}
\usepackage{lipsum}
\usepackage{enumitem}
\usepackage{stackengine}
\usepackage[font=small,labelfont=bf]{caption}
\usepackage{color}
\usepackage{adjustbox}
\usepackage{tikz}
\usepackage{tcolorbox}
\usepackage{rotating}
\usepackage{makecell}
\usepackage{colortbl}    
\usepackage{multirow}    
\usepackage{pifont}      
\usepackage{xcolor}      
\usepackage{array}       
\usepackage{amsmath}
\usepackage{amsfonts}       %
\usepackage{xspace}
\usepackage{mathtools}
\newcommand{\cmark}{\ding{51}}   
\newcommand{\xmark}{\ding{55}}   
\definecolor{oursblue}{RGB}{230,240,255} 

\newcommand{\x}{\mathbf{x}}

\definecolor{blanchedalmond}{rgb}{1.0, 0.92, 0.8}
\definecolor{carmine}{rgb}{0.59, 0.0, 0.09}
\definecolor{lightblue}{rgb}{0.22,0.45,0.70}%

\newtheorem{theorem}{Theorem}[section]

\newtheorem{proposition}[theorem]{Proposition}

\renewcommand{\mathbf}{\boldsymbol}

\makeatletter
\def\Ddots{\mathinner{\mkern1mu\raise\p@
\vbox{\kern7\p@\hbox{.}}\mkern2mu
\raise4\p@\hbox{.}\mkern2mu\raise7\p@\hbox{.}\mkern1mu}}
\makeatother

\definecolor{amaranth}{rgb}{0.9, 0.17, 0.31}
\definecolor{antiquebrass}{rgb}{0.8, 0.58, 0.46}
\definecolor{antiquefuchsia}{rgb}{0.57, 0.36, 0.51}
\definecolor{chromeyellow}{rgb}{0.31, 0.47, 0.26}

\newtcolorbox{AIbox}[2][]{aibox,title=#2,#1}
\definecolor{lightblue}{rgb}{0.22,0.45,0.70}%
\definecolor{Gray}{gray}{0.95}
\definecolor{Cornsilk}{rgb}{1.0, 0.97, 0.86}
\usepackage{enumitem}

\makeatother

\definecolor{myred}{rgb}{0.7, 0.3, 0.0}
\definecolor{myblue}{HTML}{054488}
\definecolor{mygreen}{HTML}{056b34}
\definecolor{myorange}{HTML}{ff8800}
\definecolor{mypurple}{HTML}{8400ff}
\definecolor{mypink}{HTML}{f7acb9}

\newcommand{\red}[1]{\textbf{\textcolor{myred}{#1}}}

\definecolor{myred}{rgb}{0.7, 0.3, 0.0}
\definecolor{myblue}{HTML}{054488}
\definecolor{mygreen}{HTML}{056b34}
\definecolor{tiktokpink}{HTML}{E91E63}
\definecolor{tiktokpurple}{HTML}{673AB7}
\definecolor{tiktokgray}{HTML}{9E9E9E}

\newcommand{\algo}[0]{\mbox{SimpleTIR}\xspace}

\newcommand{\mytitle}{\algo: End-to-End Reinforcement Learning for Multi-Turn Tool-Integrated Reasoning}

\usepackage{amssymb}

\title{\mytitle}

\runningtitle{\mytitle}

\author{
  Zhenghai Xue$^{1*}$,
  Longtao Zheng$^{1*}$,
  Qian Liu$^{2\dagger}$,
  Yingru Li$^2$,
  Xiaosen Zheng$^2$,
  Zejun Ma$^2$,
  Bo An$^1$
}

\affil{$^1$Nanyang Technological University, Singapore $^2$TikTok, Singapore}

\correspondingauthor{$^*$The first two authors contribute equally.}

\begin{document}

\begin{abstract}
Large Language Models (LLMs) can significantly improve their reasoning capabilities by interacting with external tools, a paradigm known as Tool-Integrated Reasoning (TIR). However, extending TIR to multi-turn scenarios using Reinforcement Learning (RL) is often hindered by training instability and performance collapse. We identify that such instability is primarily caused by a distributional drift from external tool feedback, leading to the generation of low-probability tokens. This issue compounds over successive turns, causing catastrophic gradient norm explosions that derail the training process. To address this challenge, we introduce \algo, a plug-and-play algorithm that stabilizes multi-turn TIR training. Its core strategy is to identify and filter out trajectories containing ``void turns'', i.e., turns that yield neither a code block nor a final answer. By removing these problematic trajectories from the policy update, \algo effectively blocks the harmful, high-magnitude gradients, thus stabilizing the learning dynamics. Extensive experiments show that \algo achieves state-of-the-art performance on challenging math reasoning benchmarks,  notably elevating the AIME24 score from a text-only baseline of 22.1 to  50.5 when starting from the Qwen2.5-7B base model. Furthermore, by avoiding the constraints of supervised fine-tuning, \algo encourages the model to discover diverse and sophisticated reasoning patterns, such as self-correction and cross-validation.

\vspace{0.5cm}

\coloremojicode{1F4C5} \textbf{Date}: Sep 2, 2025

\coloremojicode{1F4E7} \textbf{Correspondence}: Qian Liu~(\href{mailto:qian.liu@tiktok.com}{qian.liu@tiktok.com})

\coloremojicode{1F4BB} \textbf{Code}: \href{https://github.com/ltzheng/SimpleTIR/tree/main}{Github Repo} \qquad
\coloremojicode{1F9E0} \textbf{Model}: \href{https://huggingface.co/collections/ZhenghaiXue/simpletir-686ce09ae6e1db33b375f03d}{HuggingFace}

\end{abstract}

\maketitle
\vspace{3mm}
\begin{figure}[ht]
    \centering
    \includegraphics[width=0.90\linewidth]{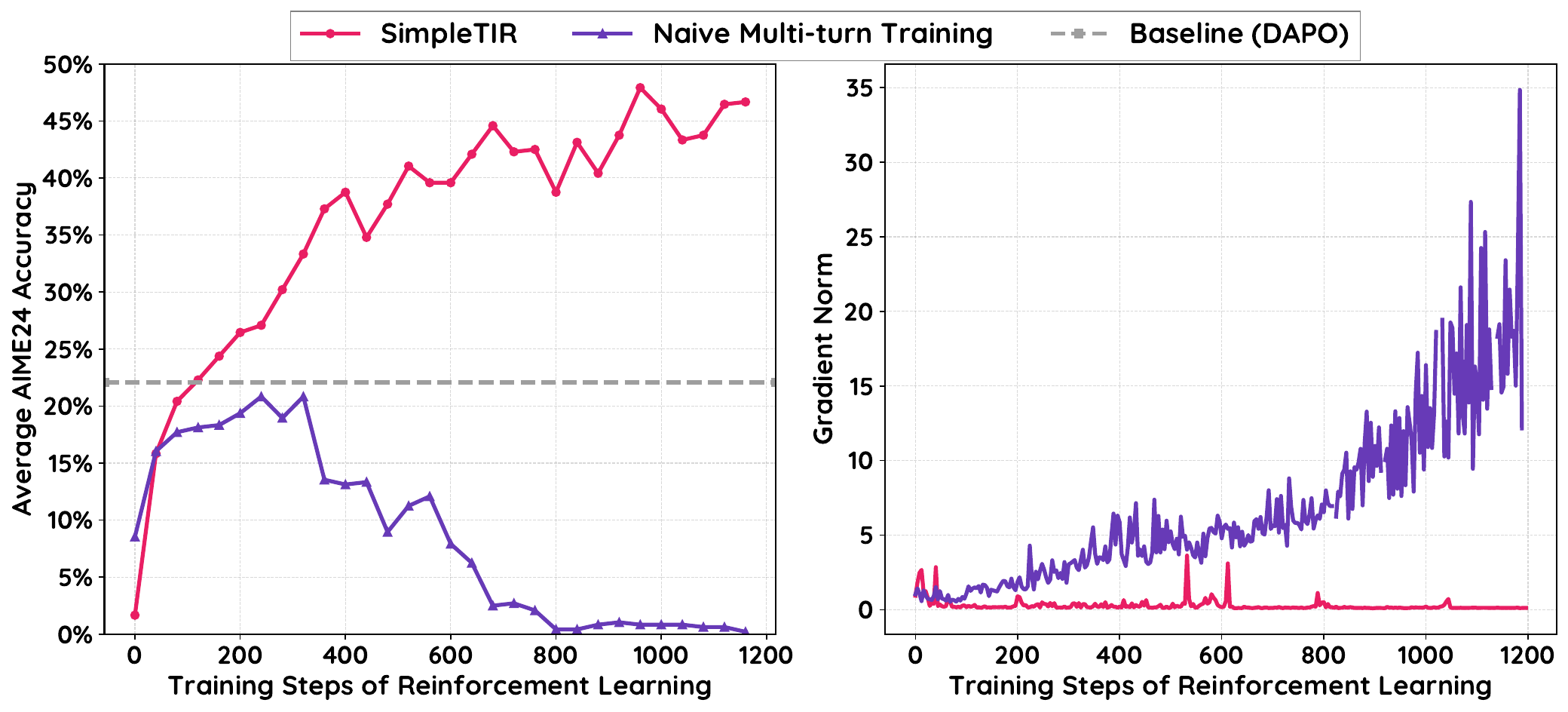}
    \caption{Starting from Qwen2.5-7B base model, The training dynamics of \textcolor{tiktokpink}{\algo} are highly stable, and it clearly outperforms the baseline method without TIR (\textcolor{tiktokgray}{DAPO}). The gradient norm remains well-behaved with almost no spikes. In contrast, \textcolor{tiktokpurple}{Naive Multi-turn Training} not only suffers from unstable dynamics and catastrophic gradient norm explosions, but also fails to match the performance of the baseline without TIR.}
    \vspace{-0.3cm}
    \label{fig:1}
\end{figure}

\section{Introduction}
Training Large Language Models (LLMs) for \emph{multi-turn} Tool-Integrated Reasoning (TIR) represents a promising frontier in Reinforcement Learning (RL). In this paradigm, LLMs iteratively reason, generate code, execute it, and utilize the output for informed reasoning in the subsequent turns. TIR addresses LLMs' inherent limitations such as poor computational accuracy and knowledge cutoffs. For example, by using a Python interpreter or a search engine, LLMs can perform precise mathematical computations or retrieve current information. Despite its clear potential, training LLMs for multi-turn TIR remains highly challenging due to frequent instability and gradient explosion issues~\citep{wang2025ragen,mai2025agent,multi-turn-kernels,Moonshot2025KimiResearcher}. One common solution is to ``cold start'' the model with Supervised Fine-Tuning (SFT) to enhance stability~\citep{feng2025retool}. However, this approach can constrain the model's discovery of novel reasoning strategies, undermining a core benefit of Zero RL training: emergent problem-solving and diverse reasoning behaviors.

In this paper, we identify a core factor contributing to this training instability: the emergence and accumulation of extremely low-probability tokens. When external tool feedback is used as model input in multi-turn TIR, such input may deviate from the model's pretrained data distribution. Although the tool feedback itself is masked when computing the policy loss~\citep{jin2025search,mai2025agent}, the model's subsequent generations inherit this distributional shift, leading to increased stochasticity. Consequently, the model is more likely to sample low-probability tokens. This issue is compounded in the multi-turn loop, as these low-probability tokens are fed back as input, exacerbating the distributional shift in subsequent turns. We make theoretical analysis of the gradient norm on softmax logits, which is part of the total gradient regarding to model parameters, and find two dominating terms with negative correlations to token probabilities. This explains the gradient explosion issues observed in prior work.

Building on this analysis, we propose \algo, an effective trajectory filtering algorithm that stabilizes multi-turn TIR training. We observe that the accumulation of low-probability tokens and high generation stochasticity frequently results in what we define as a void turn: an LLM response that contains neither a complete code block nor a final answer. Typical examples include partial code, repetitive text, or incomplete responses caused by the premature sampling of an end-of-sequence (\texttt{eos}) token. The core strategy of \algo is to filter out trajectories containing void turns. By excluding these trajectories from the policy loss computation, \algo blocks the harmful, high-magnitude gradients associated with the problematic low-probability sequences, directly addressing the gradient explosion issue. This filtering approach is also general and plug-and-play, requiring minimal modifications to be integrated into existing training frameworks for improved stability and performance with almost no extra cost.

To demonstrate the effectiveness of \algo, we conduct comprehensive experiments on challenging mathematical reasoning tasks. When applied to the Qwen2.5-7B base model, \algo achieves state-of-the-art performance in multi-turn TIR, improving the AIME24 score from a text-only baseline of 22.1 to 50.5. Our ablation studies confirm that filtering trajectories with void turns is the crucial component for stabilizing training, overcoming the instability that plagues naive multi-turn approaches and enabling significant performance gains. Finally, we highlight a key advantage of our Zero RL approach. In contrast to methods that rely on a ``cold-start'' SFT phase, \algo encourages the model to discover novel and diverse reasoning patterns, such as cross-validation, progressive reasoning, and self-correction.

\label{sec:intro}

\section{Preliminaries}

End-to-end training of multi-turn Tool-Integrated Reasoning (TIR) agents with Reinforcement Learning (RL) is challenging due to its compositional structure. We therefore model the process as a \textbf{Hierarchical Markov Decision Process}~\citep{hauskrecht1998hierarchical}, which separates decision-making into two levels: a high-level policy governing the sequence of conversational turns and a low-level policy for generating tokens within each turn.

\subsection{Hierarchical MDP Formulation for Multi-Turn TIR}

A full interaction trajectory is a sequence $o = (q, l_0, f_0, \dots, l_{K-1}, f_{K-1})$, where $l_k$ is the model's generated response at turn $k$ and $f_k$ is the subsequent tool feedback.

The \textbf{high-level MDP}, $\mathcal{M}_H = \langle \mathcal{S}_H, \mathcal{A}_H, T_H, R_H, \gamma_H \rangle$, operates at the turn level to govern the overall strategy.
\begin{itemize}[leftmargin=*]
    \item \textbf{State ($S_k$)}: $S_k = (q, l_0, f_0, \dots, l_{k-1}, f_{k-1})$. This represents the complete conversation history before the current turn.
    \item \textbf{Action ($L_k$)}: An option, or high-level sub-policy, that generates the entire response $l_k$ for the current turn.
    \item \textbf{Transition ($T_H$)}: $S_{k+1} = S_k \circ (l_k, f_k)$, where the state evolves by appending the generated response and its corresponding tool feedback.
    \item \textbf{Reward ($R_H$)}: $R(o)$, a terminal reward assigned to the final trajectory based on its overall success.
\end{itemize}

The \textbf{low-level MDP}, $\mathcal{M}_L = \langle \mathcal{S}_L, \mathcal{A}_L, T_L, R_L, \gamma_L \rangle$, operates at the token level to execute the chosen high-level action~\citep{li2024remax}.
\begin{itemize}[leftmargin=*]
    \item \textbf{State ($s_t$)}: $s_t = S_k \circ (a_1, \dots, a_{t-1})$. This is the sequence of tokens generated so far within the current turn $k$.
    \item \textbf{Action ($a_t$)}: $a_t \in \mathcal{A}$, a single token selected from the model's vocabulary.
    \item \textbf{Transition ($T_L$)}: $s_{t+1} = s_t \circ a_t$, where the state evolves by deterministically appending the selected token.
    \item \textbf{Reward ($R_L$)}: $R_L = 0$. The low-level policy receives no intrinsic reward, as its only goal is to complete the high-level action.
\end{itemize}
In this framework, $\circ$ denotes concatenation, and we set the discount factor $\gamma = \gamma_H = \gamma_L =1$. We train a single, unified policy $\pi_\theta(a_t | s_t)$ to implicitly solve this two-level problem.

\subsection{Joint Policy Optimization and Feedback Masking}
\label{sec:policy_opt}

The unified policy $\pi_\theta$ is trained using {Group Relative Policy Optimization (GRPO)}~\citep{deepseek-ai-2025}. GRPO circumvents the need for a learned value function~\citep{li2024remax} by calculating the advantage based on the relative performance within a group of $G$ trajectories sampled from the same prompt. The advantage for trajectory $o_i$ is \(\hat{A}_{i} = \frac{r_i - \operatorname{mean}\left(\{r_j\}_{j=1}^G\right)}{F_{\text{norm}}\left(\{r_j\}_{j=1}^G\right)}\),
where $r_i$ is the terminal reward from the high-level MDP.

A critical adaptation is required for the TIR setting. The policy is only responsible for generating response tokens ($l_k$), not the environment-provided feedback tokens ($f_k$). To ensure correct credit assignment, we employ \textbf{feedback token masking}~\citep{jin2025search,mai2025agent}. The loss is accumulated only over the timesteps corresponding to the agent's actions, effectively excluding feedback tokens from the gradient computation.

This leads to our final training objective, $\mathcal{J}_{\text{TIR}}(\theta)$:
\begin{equation}
\mathcal{J}_{\text{TIR}}(\theta) = \mathbb{E}_{\substack{q \sim q_0, \\ \{o_i\}_{i=1}^G \sim \pi_{\theta_{\text{old}}}(\cdot|q)}} \left[ \frac{1}{G} \sum_{i=1}^G \frac{1}{\sum_t m_{i,t}} \sum_{t=1}^{|o_i|} m_{i,t} \cdot L_{\text{CLIP}}(\theta, i, t) \right],
\end{equation}
where $m_{i,t}$ is a binary mask that is $1$ if the token at step $t$ belongs to any response $l_k$ and $0$ otherwise. The term $L_{\text{CLIP}}$ is the standard clipped surrogate objective from PPO~\citep{schulman2017proximal,wang2019divergence}:
\begin{equation}
    L_{\text{CLIP}}(\theta, i, t) = \min \left(\rho_{i, t}(\theta) \hat{A}_{i}, \operatorname{clip}(\rho_{i, t}(\theta), 1-\varepsilon, 1+\varepsilon) \hat{A}_{i}\right),
\end{equation}
with the importance sampling ratio $\rho_{i, t}(\theta) = \frac{\pi_{\theta}(o_{i, t} \mid o_{i, <t})}{\pi_{\theta_{\text{old}}}(o_{i, t} \mid o_{i, <t})}$.

\section{Methodology}

We first diagnose a core source of instability in multi-turn TIR: the emergence of low-probability tokens. We demonstrate how these tokens drive gradient explosions and create a credit assignment dilemma during Zero-RL training. Building on this analysis, we introduce \textsc{SimpleTIR}, a simple yet effective trajectory filtering method that stabilizes training while encouraging the model to develop sophisticated, multi-turn reasoning strategies.

\subsection{The Emergence of Low-Probability Tokens in Multi-Turn TIR}
\label{sec:multi_turn_issue}

\begin{figure}[t]
    \centering
    \includegraphics[width=\linewidth]{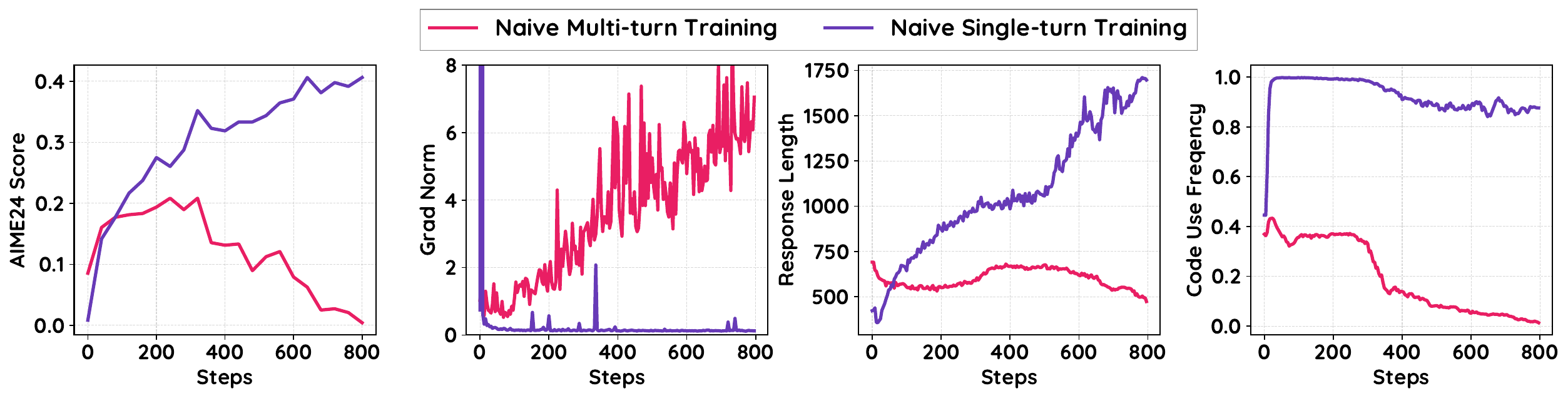}
    \vspace{-0.2cm}
    \caption{Training statistics comparing naive \textcolor{tiktokpurple}{single-turn} and \textcolor{tiktokpink}{multi-turn} TIR. Single-turn training proceeds smoothly and achieves higher performance, while multi-turn training is unstable.}
    \vspace{-0.3cm}
    \label{fig:stable_single_turn}
\end{figure}

The instability of multi-turn TIR training is a known challenge~\citep{wang2025ragen,mai2025agent}. To isolate the cause, we contrast it with a minimal, single-turn TIR setting where the model produces exactly one response containing reasoning and an optional code block. As shown in Fig.~\ref{fig:stable_single_turn}, the \textcolor{tiktokpurple}{single-turn} baseline trains smoothly and achieves reasonable performance, whereas the \textcolor{tiktokpink}{multi-turn} equivalent suffers from performance collapse and recurring gradient spikes.

The key difference lies in the feedback loop. In multi-turn TIR, the tool feedback $f_k$ from turn $k$ is concatenated into the prompt for turn $k+1$. Because this feedback originates from an external interpreter, it can deviate significantly from the LLM's learned data distribution. Conditioned on such out-of-distribution (OOD) input, the model's subsequent generations can drift away from pretrained patterns, becoming highly stochastic and assigning unnaturally low probabilities to selected tokens.

\begin{figure}[t]
    \centering
    \includegraphics[width=0.95\linewidth]{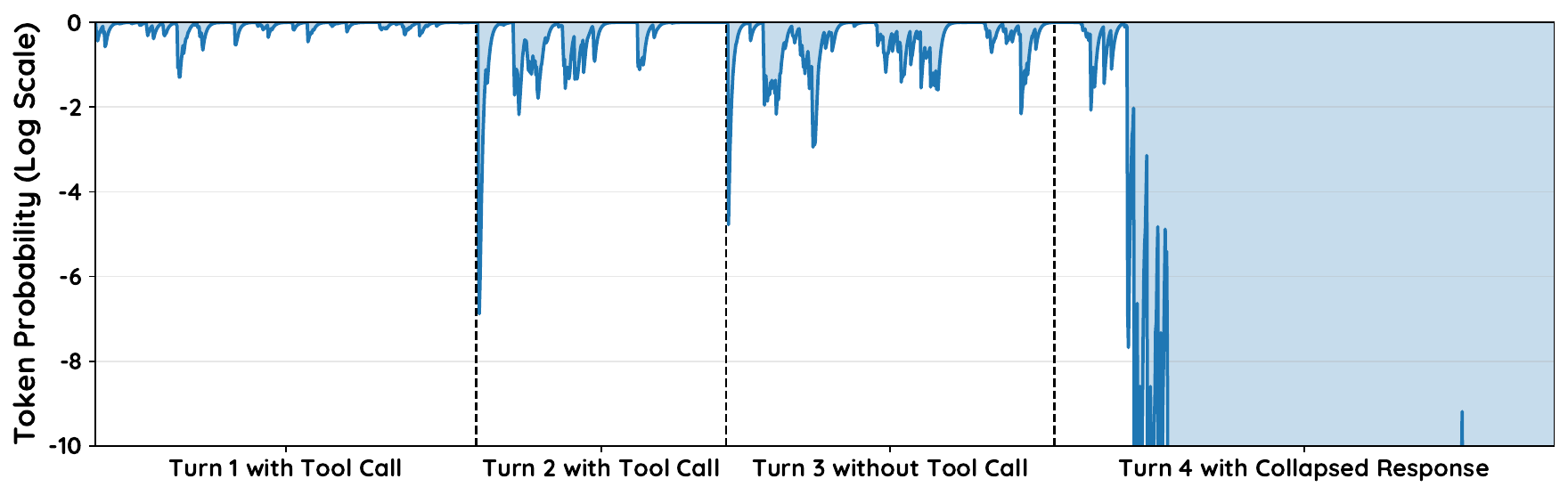}
    \caption{Visualization of token probabilities in a multi-turn TIR trajectory. The y-axis is log-scaled. Distributional drift from tool feedback in early turns leads to a collapse in token probabilities in later turns.}
    \label{fig:token_prob_vis}
\end{figure}

We verify this phenomenon with a case study in Fig.~\ref{fig:token_prob_vis}. The tool feedback in the early turns contains tokens with extremely low probabilities, confirming their OOD nature. While masking the loss on these feedback tokens is a known practice~\citep{jin2025search,mai2025agent}, it is insufficient. The distributional drift it induces contaminates subsequent model generations. As seen in Fig.~\ref{fig:token_prob_vis}, while token probabilities in Turn 1 are high, low-probability segments emerge in the model's own text in Turns 2 and 3. This compounding drift culminates in a collapsed, nonsensical response with extremely low token probabilities in Turn 4.

\subsection{How Low-Probability Tokens Compromise Zero-RL Training}

Having established the emergence of low-probability tokens, we now analyze their two primary detrimental effects on Zero-RL training: gradient explosion and misaligned credit assignment.

\paragraph{Gradient Explosion.} A primary failure mode in multi-turn TIR is the explosion of gradient norms (Fig.~\ref{fig:stable_single_turn}). To formalize this, we analyze the policy gradient with respect to the pre-softmax logits $\mathbf{z}$~\citep{li2025logit}.

\begin{proposition}
\label{prop:grad_norm}
Consider a token $c$ at timestep $t$ of a trajectory $o_i$. The L2 norm of the policy gradient with respect to the logits $\mathbf{z}_t$ is:
\begin{equation}
    \|\nabla_{\mathbf{z}_t} \mathcal{J}_{\text{TIR}}\|_2 = \frac{m_{i,t}}{\sum_j m_{i,j}} \cdot \rho_{i,t}(\theta) \cdot g_{i,t} \cdot |\hat{A}_i| \cdot \sqrt{1 - 2P(c) + \sum_{j \in \mathcal{A}} P(j)^2},
\end{equation}
where $m_{i,t}$ is the feedback mask, $\rho_{i,t}(\theta)$ is the importance ratio, $|\hat{A}_i|$ is the absolute advantage, $P$ is the policy's probability distribution $\pi_\theta(\cdot|o_{i,<t})$, and $g_{i,t}$ is a gating function active when the PPO update is not clipped.
\end{proposition}

Proposition~\ref{prop:grad_norm} reveals that the gradient norm is highly sensitive to two factors that are exacerbated by low-probability tokens:
\begin{itemize}[leftmargin=*]
    \item \textbf{Unclipped Importance Ratio:} The ratio $\rho_{i, t}(\theta) = \frac{\pi_{\theta}(o_{i, t} \mid o_{i, <t})}{\pi_{\theta_{\text{old}}}(o_{i, t} \mid o_{i, <t})}$ is a primary source of gradient spikes. For a negatively-rewarded trajectory ($\hat{A}_i < 0$), this ratio is unbounded from above. If token $c$ was generated with a very low probability by the old policy, $\pi_{\theta_{\text{old}}}(c|\cdot)$ is minute. Even a small update to $\pi_\theta(c|\cdot)$ can cause $\rho_{i,t}(\theta)$ to explode, leading to the gradient spikes observed in training.
    \item \textbf{Sustained High Gradient Norm:} The probability-dependent term, $\sqrt{1 - 2P(c) + \sum_j P(j)^2}$, can sustain large gradients. When the policy assigns a low probability to the sampled token $c$, $1-2P(c)$ nears its maximum of 1. If the policy is otherwise confident (i.e., the distribution is sharp), the collision probability $\sum_j P(j)^2$ remains large, preventing the gradient norm from diminishing and thus contributing to unstable training~\citep{li2025logit}.
\end{itemize}

\paragraph{Misaligned Credit Assignment.} Beyond gradient instability, low-probability tokens introduce a severe credit assignment problem. As seen in Fig.~\ref{fig:token_prob_vis}, these tokens are more prevalent in later turns. With a sparse, terminal reward, a trajectory that fails in its final turns receives a single negative reward for the entire sequence. This signal does not distinguish between correct, high-probability reasoning in early turns and the faulty, low-probability tokens that caused the eventual failure. This dynamic unfairly penalizes valid multi-turn behavior, causing the policy to collapse toward safer, single-turn generations.

\subsection{SimpleTIR: Stabilizing Training by Filtering Void Turns}
\label{sec:multi_turn_method}

Given that low-probability tokens are the root cause, simple heuristics like masking high-perplexity trajectories or clipping the importance ratio may seem appealing. While effective in some contexts~\citep{chujie2025group,zhang2025r1}, we show in Fig.~\ref{fig:curve} (bottom) that these methods fail to resolve instability in multi-turn TIR, as their thresholds are difficult to tune and they do not solve the credit assignment problem.

\begin{figure}[t]
    \centering
    \includegraphics[width=\linewidth]{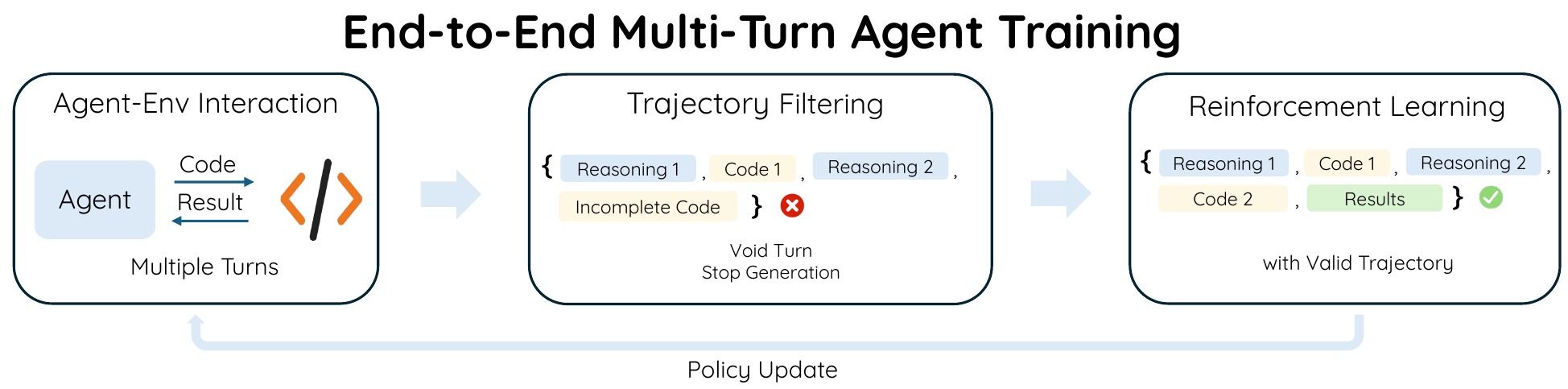}
    \caption{An overview of \textsc{SimpleTIR}. During the policy update, \textsc{SimpleTIR} identifies and filters out entire trajectories that contain a \textit{void turn}—an LLM response that fails to produce either a complete code block or a final answer.}
    \label{fig:algo}
\end{figure}

A more robust filtering criterion is needed. We observe that the collapsed Turn 4 in Fig.~\ref{fig:token_prob_vis} follows a turn that produced neither a tool call nor a final answer. Intuitively, such a turn makes no progress in the reasoning process. We define these as \textbf{void turns}. A void turn is often a symptom of distributional drift, where high generation stochasticity leads to a premature end-of-sequence token. Because void turns are rare in successful trajectories but indicative of abnormal ones, they serve as a powerful heuristic for identifying problematic trajectories.

This insight leads to the \textsc{SimpleTIR}{} algorithm, illustrated in Fig.~\ref{fig:algo}. The procedure is simple: for each sampled trajectory, we inspect its turns. If any turn contains neither a complete code block nor a final answer, it is labeled a void turn. We then \textbf{mask the policy loss for the entire trajectory}, removing it from the batch before the GRPO update. This single step simultaneously prevents the large gradients from low-probability tokens from backpropagating and corrects misaligned credit assignment by ensuring successful early turns are not penalized for a later collapse. \textsc{SimpleTIR}{}'s filtering approach is agnostic to the specific RL algorithm used and is orthogonal to other recent improvements in RL for LLM reasoning~\citep{chujie2025group,yao2025offpolicy,chen2025minimax}.

\subsection{Implementation Details}

To further enhance training stability and efficiency, we adopt several key practices. First, to avoid out-of-distribution special tokens when using base models, we do not use chat templates. Instead, we prepend tool outputs with a simple prefix, ``Code Execution Result:''. Second, to provide a shortcut for simple tasks and improve sample efficiency, we prepend every LLM-generated code block with a `final\_answer` function, allowing the model to terminate and answer within a single turn if possible. Finally, to prevent the model from hallucinating tool outputs, we strictly stop LLM generation after a complete code block is formed and always append the true, external tool feedback before the next turn begins.

\section{Experiments}
\newcommand{\ncolumn}{9}
\begin{table}[t]
    \centering
    \footnotesize
    \setlength{\tabcolsep}{6.5pt}     
    \renewcommand{\arraystretch}{1.15} 
    \caption{Performance comparison on various math benchmarks. Check and cross marks in the ``TIR'' column refers to whether the method involves TIR during training and evaluation. Slash, check, and cross marks in the ``Zero RL'' column refers to whether the model is untrained, trained with the Zero RL setting, or trained with other settings. The ``From'' column indicates the type of  We fill the scores with - if they are not provided in respective reports.}
    \vspace{0.2cm}
    \label{tab:math_results}
    \begin{tabular}{@{}lcccrrrrrr@{}}
        \toprule
        \textbf{Model} &
        \textbf{TIR} &
        \textbf{Zero RL} &
        \textbf{From} &
        \textbf{AIME24} & \textbf{AIME25} & \textbf{MATH500} &
        \textbf{Olympiad} & \textbf{AMC23} & \textbf{Hmmt 25} \\
        \midrule 
        \multicolumn{\ncolumn}{c}{\itshape Models based on Qwen2.5-7B} \\
        \midrule
        Qwen2.5-7B        & \xmark & / & Base &  3.2  & 1.1 & 51.9 & 15.4 & 21.7 & 0.0 \\
        Qwen2.5-7B-TIR    & \cmark & / & Base &  1.7 & 0.6 & 18.0 & 6.2 & 10.8 & 1.9 \\
        SimpleRL-Zoo-7B     & \xmark  & \cmark & Base & 15.6 & -  & 78.2 & 40.4 & 62.5 & - \\
        ToRL-7B  & \cmark & \xmark & Math-Inst & 40.2 & 27.9 & 82.2  & 49.9 & 75.0 & - \\
        Effective TIR-7B & \cmark & \xmark & Math & 42.3 & 29.2  & 86.4 & - & 74.2 & -\\
        ARPO-7B & \cmark & \xmark & Inst & 30.0 & 30.0 & 78.8 & - & - & - \\
        ZeroTIR-7B        & \cmark & \cmark & Base & 39.6 & 25.0 & 80.2 & - & - & 22.5 \\
        \rowcolor{oursblue}
        \algo-7B & \cmark & \cmark & Base & \textbf{50.5} & \textbf{30.9} & \textbf{88.4} & \textbf{54.8} & \textbf{79.1} & \textbf{29.7} \\
        \midrule
        \multicolumn{\ncolumn}{c}{\itshape Models based on Qwen2.5-32B} \\        
        \midrule
        Qwen2.5-32B        & \xmark & / & Base &  4.2 & 1.6 & 43.1 & 17.8 & 28.0 & 0.2 \\
        Qwen2.5-32B-TIR    & \cmark & / & Base &  7.1 & 5.0 & 37.0 & 16.9 & 20.0 & 5.2 \\
        DAPO               & \xmark & \cmark & Base & 50.0 & - & - & - & - \\
        ReTool             & \cmark & \xmark & Math-Inst & 67.0 & 49.3 & - & - & - & - \\
        ZeroTIR-32B        & \cmark & \cmark & Base & 48 & 27 & 87.8 & - & - & 20.0 \\
        \rowcolor{oursblue}
        \algo-32B & \cmark & \cmark & Base & 59.9 & 49.2 & 92.9 & 63.7 & 91.6 & 34.6 \\
        \bottomrule
    \end{tabular}
\end{table}

\subsection{Setup}
\paragraph{Training} We prepare our training code with the VeRL~\citep{sheng2024hybridflow} and Search-R1~\citep{jin2025search} framework. We use Sandbox Fusion as an asynchronous code interpreter. The training datasets are Math3-5 from SimpleRL~\citep{zeng2025simplerl} and Deepscaler~\citep{deepscaler2025}. \algo follows the Zero RL setting and uses the unaligned Qwen-2.5 series as the base models, including Qwen-2.5-7B and Qwen-2.5-32B. During training, the rollout batch size is set to 512, and the mini update size is set to 128. The maximum response length is initially set to 16K, with a maximum of five turns of code execution. When the average response length plateaus, we increase the maximum response length to 24K and the largest number of turns to 10.
Other training hyperparameters are in Appendix~\ref{sec:hyper}.
\paragraph{Evaluation}
Our evaluation is conducted on Math500~\citep{hendrycks2021measuring}, AIME24, AIME25, AMC23, and Hmmt Feb 25, using a temperature of 1 and reporting average@32 scores to reduce variance, following~\citet{yu2025dapo}. For comparison, we consider three categories of baselines. The first is non-TIR Zero RL, where we use SimpleRL-Zoo~\citep{zeng2025simplerl} and DAPO~\citep{yu2025dapo} as representative baselines. The performance gap between these methods and \algo highlights the advantage of incorporating TIR in mathematical reasoning. The second category is TIR RL from cold-start or specialized models, which includes ReTool~\citep{feng2025retool}, collecting cold-start datasets for supervised finetuning on Qwen2.5-Math-32B-Instruct, ARPO~\citep{dong2025agentic}, finetuning Qwen2.5-7B-Instruct, as well as ToRL~\citep{li2025torl} and Effective CIR~\citep{bai2025towards}, both applying RL to the Qwen2.5-Math series. The final category is Zero RL with TIR, where, to the best of our knowledge, Zero-TIR~\citep{mai2025agent} is the only method that strictly follows the Zero RL paradigm by training TIR models directly from base models.
starting from unaligned base models when training TIR models.
\begin{figure}[t]
    \caption{\textbf{Top}: Training curves for \algo with different maximum number of turns. \textcolor{tiktokpink}{\algo with maximum 10 turns} is resumed at 200 steps from \textcolor{tiktokpurple}{\algo with maximum 5 turns}. \algo clearly benefits from scaling interaction turns from 1 to 5. \textbf{Bottom}: The training curves for ablation studies in the first 320 steps. Trajectory filtering with \textcolor{tiktokgray}{high importance ratios} or \textcolor{tiktokpurple}{low probability tokens} cannot resolve the challenge of training instability, while \textcolor{tiktokpink}{\algo} suffers less from low probability tokens and gradient explosion.} 
    \vspace{0.3cm}
    \label{fig:curve}
    \centering
    \includegraphics[width=0.999\linewidth]{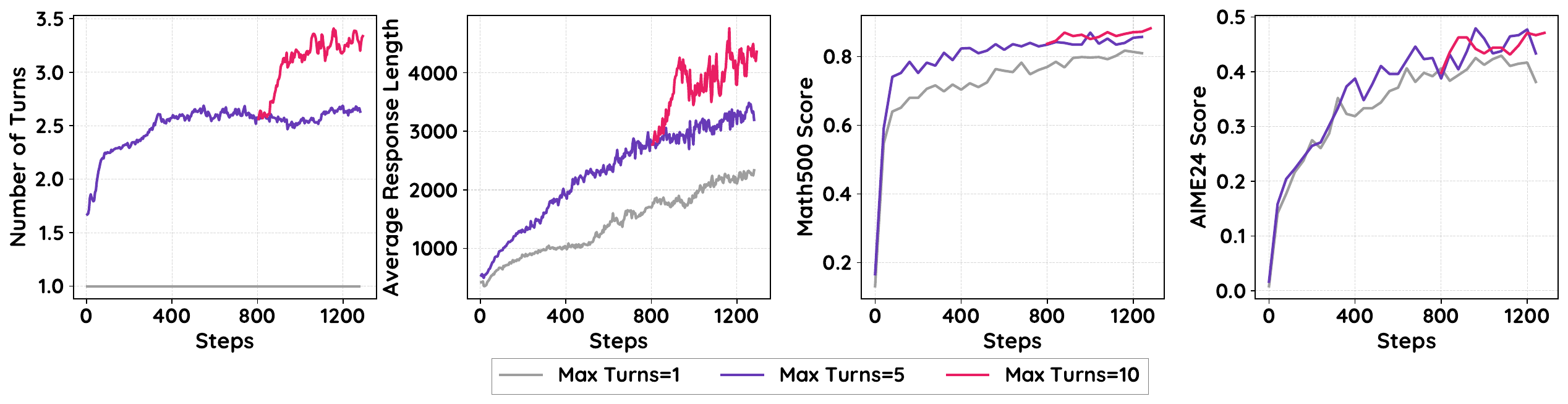}
    \includegraphics[width=0.999\linewidth]{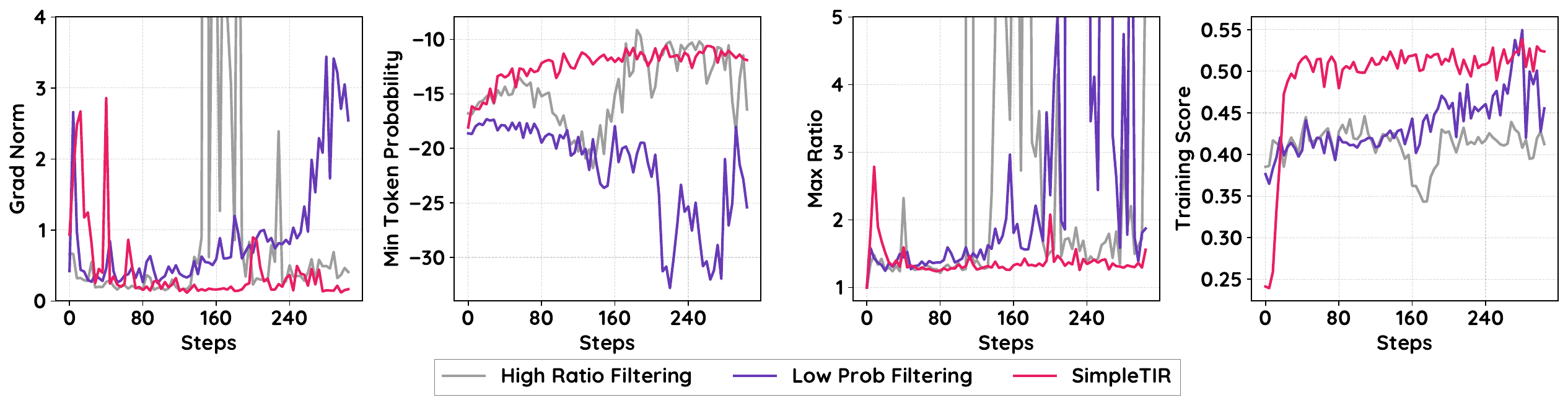}
    \vspace{-0.5cm}
\end{figure}
\begin{table}[t]
    \centering
    \caption{Results of ablation studies. Considering the unstable training of ablated methods, we report the highest scores within 1000 gradient steps. ``Naive Multi-Turn'' directly applies RLVR in multi-turn TIR. ``Low Prob'' and ``High Ratio'' filtering refers to masking the policy loss on tokens with lowest probabilities or highest importance ratio.}
    \vspace{0.2cm}
    \label{tab:ablation}
    {\fontsize{9.0}{11}\selectfont
    \begin{tabular}{cccccc}
    \toprule
         & \algo-7B & Naive Multi-Turn & Low Prob Filtering & High Ratio Filtering & Stop Generation w/o Filtering \\
         \midrule
        AIME24 & 50.5 & 20.8 & 23.3 & 26.3 & 26.1\\
        Math500 & 88.4 & 73.1 & 72.8 & 75.0 & 77.3\\
        \bottomrule
    \end{tabular}
    }
\end{table}
\subsection{Training Results}
The training results are listed in Tab.~\ref{tab:math_results}. \algo demonstrates significant performance improvement over base models and outperform all baselines of Zero RL, either with or without TIR. \algo can also outperform baselines starting from Qwen2.5-Math-7B series, such as ToRL and Effective TIR.
Comparing with methods not following Zero RL, it is shown that cold start significantly boosts performance, with ReTool-32B obtaining the highest scores on AIME24 and AIME25. The advantage of Zero RL over cold start lies in the diversity of reasoning patterns, as discussed in Sec.~\ref{sec:analysis}.

\subsection{Training Curves and Ablation Studies}

We show the training curve of \algo with 1, 5, and 10 turns of generation in Fig.~\ref{fig:curve} (Top). In all these settings, \algo exhibits constant and smooth increases of the average response length and performance scores. The average number of turns first arises quickly then remains constant for multi-turn \algo. We also observe that the response length and the Math500 score scales with more turns, while the AIME24 score does not benefit clearly. This indicates that different tasks require distinct reasoning patterns. Some may be solvable with few steps of reasoning, but others will take a number of external feedback before reaching the correct answer. 

We also conduct ablation studies to demonstrate the effectiveness of trajectory filtering in \algo. We first investigate two alternative filtering criteria: high importance ratio and low token probabilities, as specified in the first paragraph of Sec.~\ref{sec:multi_turn_method}. As shown in Fig.~\ref{fig:curve} (Bottom), these two filtering approach cannot resolve the issue of gradient explosion, exhibiting unstable curves of training scores. \algo features a more stable curve of gradient norm, thanks to the mild token probability distributions. This demonstrates the effectiveness of void turn filtering in stabilizing multi-turn TIR training. We then consider an ablation method where LLM generation is terminated on void turns but resulting trajectories are not filtered when computing policy loss. According to the validation results in Tab.~\ref{tab:ablation}, this method is also inferior to \algo. This can be attributed to misaligned credit assignment since trajectories containing void turns can hardly obtain positive outcome. \algo handles such issue by masking the loss of whole responses containing void turns.

\begin{figure}[t]
    \caption{Demonstration of three reasoning patterns observed in responses generated by \algo.}
    \label{fig:pattern}
    \centering
    \includegraphics[width=0.95\linewidth]{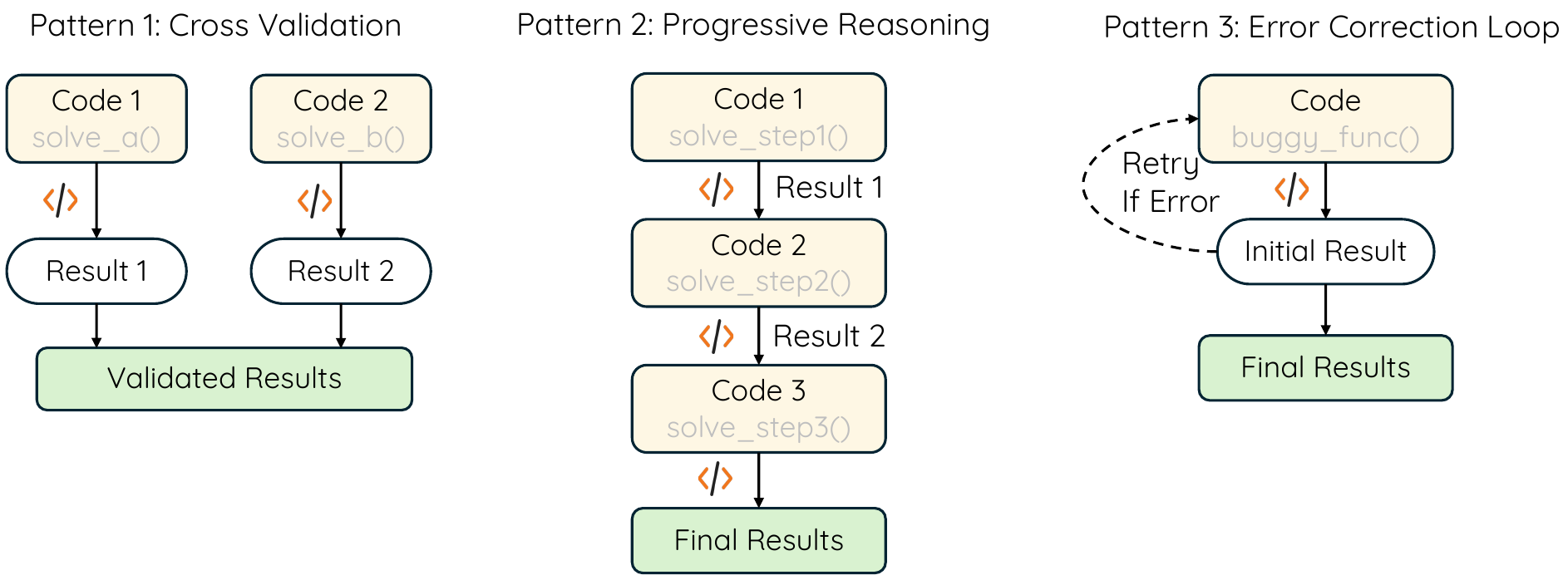}
\end{figure}
\begin{table}[t]
\footnotesize
\caption{Comparison of reasoning pattern frequencies in ReTool and \algo-32B responses. The summation of frequencies may exceed 100\% as there may be more than one reasoning patterns in one response.}
\vspace{0.3cm}
\centering
\setlength{\tabcolsep}{8pt}     
\begin{tabular}{llll}
\toprule
& \makecell[l]{Progressive\\Reasoning (\%)} & \makecell[l]{Cross\\ Verification (\%)} & \makecell[l]{Error\\Correction (\%)} \\
\midrule 
ReTool & 18.9 & 82.4 & 25.8 \\
\midrule 
\algo-32B & 46.5 & 86.0 & 38.0 \\
\bottomrule
\end{tabular}
\end{table}
\subsection{Emergence of Diverse Reasoning Behaviors}
\label{sec:analysis}
Thanks to the framework of Zero RL training, \algo automatically reinforces useful reasoning patterns obtained in the pretraining phase, rather than sticking to predefined patterns in the SFT dataset. In Appendix~\ref{sec:append_reasoning_behavior}, we show \algo responses with diverse multi-turn reasoning behaviors. They are mostly combinations of the three main reasoning patterns illustrated in Figure~\ref{fig:pattern}, namely Cross Validation, Progressive Reasoning, and Error Correction.

We also use Claude-3.7-Sonnet to identify and count the frequency of reasoning patterns in responses generated by ReTool and \algo-32B.
The responses are filtered so that they all lead to the correct final answer. Both models demonstrate a strong tendency to conduct multiple rounds of cross verification. Meanwhile, \algo-32B exhibits more instances of progressive reasoning and error correction. This illustrates the advantage of Zero RL, which preserves more diversity in reasoning patterns.

\section{Related Work}
\label{sec:related_work}

\subsection{Zero RL for LLM Reasoning}
DeepSeek-R1~\citep{deepseek-ai-2025} first shows that starting from an unaligned base model, large-scale RL training with outcome reward can unlock emergent chain-of-thought reasoning ability. Such paradigm is later referred to as Zero RL.
SimpleRL~\citep{zeng2025simplerl} provides a reproducible cookbook to run Zero RL on various open-source base models. Open-Reasoner-Zero~\citep{hu2025openreasoner} proposes that vanilla PPO with GAE ($\lambda=1, \gamma=1$) without KL regularization is sufficient to scale up Zero RL training. DAPO~\citep{yu2025dapo} introduce several training details that makes Zero RL training stable and efficient, such as raising the high clip ratio of PPO and GRPO and filtering tasks with 0 or 100\% solve rate. Dr. GRPO~\citep{liu2025understanding} proposes to remove the length normalization term. \algo also follows the Zero RL pipeline and is orthogonal to training algorithms for Zero RL without TIR.

\subsection{RL for Tool Integrated Reasoning}
Several recent works focus on applying RL to improving the tool use ability of LLMs. Search-R1~\citep{jin2025search} and R1-Search~\citep{song2025r1} focus on question-answering tasks, utilizing the search tool. For mathematical reasoning tasks, python interpreter can be a useful tool to conduct numerical calculations or enumerations. ReTool~\citep{feng2025retool} employs a cold-start SFT phase before RL. ToRL~\citep{li2025torl} and Effective CIR~\citep{bai2025towards} explore training recipes on math-specialized bases. These pipelines often rely on domain data, instruction tuning, or other supervision that introduce bias and complexity; in contrast, Zero RL is more general yet notoriously unstable in multi-turn settings. Our work directly addresses this stability gap under Zero RL by filtering trajectories with void turns. ZeroTIR~\citep{mai2025agent} is also explicitly framed in the Zero RL setting. It proposes several stabilizing techniques that are orthogonal to our approach. There is also a theoretical explanation~\citep{lin2025understanding} on why TIR is more effective than text-only reasoning. \algo serves as a good empirical evidence of their claim.

We leave related work on stabilizing RL training in Appendix~\ref{append:related}.
\section{Conclusion}
In this work, we introduce \algo, an RL framework designed to stabilize and enhance multi-turn TIR under the Zero RL setting. By addressing the key challenge of harmful negative samples via filtering out trajectories with void turns, our method achieves stable training dynamics and improves reasoning performance across a variety of mathematical benchmarks. Beyond state-of-the-art results starting from the Qwen2.5-7B model, \algo also encourages the emergence of diverse reasoning patterns. These results highlight the potential of end-to-end multi-turn TIR RL, without relying on cold-start human data, as a pathway to scalable and reliable multi-turn reasoning in future LLM agent development.
\paragraph{Limitations and Future Work}
While effective, our method has several limitations. First, we use void turns as an indicator of low-probability tokens in multi-turn TIR. However, this indicator may not be directly applicable to tasks beyond multi-turn TIR. Second, we currently restrict the maximum number of turns to 10 for mathematical reasoning, though more interactions may be required for complex multi-turn agent tasks. Third, our training relies on a highly parallel sandbox for code execution. Therefore, the development of a faster and more reliable sandbox is an important direction for future work. Finally, achieving fully asynchronous rollout and reward calculation remains an open challenge. These limitations raise additional concerns around rollout efficiency, memory management, and credit assignment, which we leave for future exploration.
\label{sec:conclusion}

\paragraph{Acknowledgement}
We thank Lang Feng and Jiacheng Xu for helpful discussions.
\clearpage
\bibliography{main}

\appendix
\newpage
\appendix
\section{Extended Related Work}
\label{append:related}
\subsection{Stabilizing RL Training}
Training instability is a significant challenge when applying RL to LLMs, often manifesting as entropy collapse and gradient norm explosions. 
Entropy-based methods explicitly maintain policy entropy or encourage re-generation at pivotal tokens to delay distributional narrowing~\citep{cui2025entropy,li2025cure,liu2025prorl}. 
Recent methods control the importance sampling ratio to reduce gradient variance and brittle updates by reweighting or constraining likelihood ratios, e.g., from token-level IS to sequence-level objectives and clipping~\citep{yao2025offpolicy,chen2025minimax,zhao2025geometric,zheng2025group}. 
Data and trajectory filtering stabilizes training by discarding uninformative or harmful samples, e.g., multi-sample-then-filter schemes~\citep{shrivastava2025sample}. 
From the perspective of the learning signal itself, negative-only gradient updates have been shown to improve stability and generalization without sacrificing exploration, and more generally to focus updates on low-probability/high-entropy branching tokens~\citep{zhu2025surprising}. 
\algo departs from the above methods by targeting the root cause specific to TIR, i.e., distribution shift induced by external tool outputs compounded by multi-turn error accumulation. 
It is also orthogonal to entropy regularization, IS ratio control, and negative-gradient schemes. 

\section{Example Responses}
\subsection{Incomplete Response}
\label{sec:append_incomplete}
\begin{table*}[!t]
    \centering
    \caption{An example from \algo-7B on AIME24 dataset. Responses generated in different turns are separated with a vertical line. Response Part-II contains a void turn, where no complete code blocks nor boxed answers are generated. Both in and after the void turn, the model generates corrupted responses.}
    \label{tab:append_incomplete}
    \fontsize{9pt}{11pt}\selectfont
    \begin{tabular}{p{0.98\linewidth}}
    \midrule
        \rowcolor{gray!20}\textbf{Example \#1 from AIME24 Dataset} \\
    \midrule
        \textbf{Question:} \\
Eight circles of radius $34$ are sequentially tangent, and two of the circles are tangent to $AB$ and $BC$ of triangle $ABC$, respectively. $2024$ circles of radius $1$ can be arranged in the same manner. The inradius of triangle $ABC$ can be expressed as $\frac{m}{n}$, where $m$ and $n$ are relatively prime positive integers. Find $m+n$.

\begin{asy}
size(8cm);
pair A = (2,1);
pair B = (0,0);
pair C = (3,0);
dot(A^^B^^C);
label("$A$", A, N);
label("$B$", B, S);
label("$C$", C, S);
draw(A--B--C--cycle);
for(real i=0.62; i<2.7; i+=0.29){
  draw(circle((i,0.145), 0.145));
}
\end{asy}
\\
        \textbf{Labeled Answer:} \\
        197 \\
    \midrule
        \rowcolor{gray!20}
        \textbf{Model Output by \algo-7B (Part-I)}\\
    \midrule
To solve this problem, we need to determine the inradius of triangle \(ABC\) given the arrangement of circles. Here is the step-by-step reasoning:

**Understand the Geometry:**

- We have eight circles of radius \(34\) that are tangent to each other sequentially, with two of them being tangent to the sides \(AB\) and \(BC\) of triangle \(ABC\).
- Additionally, \(2024\) circles of radius \(1\) can fit in the same manner.

**Determine the Distance Between Centers of the Larger Circles:**

- The centers of the eight tangent circles of radius \(34\) will form a chain where each circle is tangent to the next, meaning the distance between the centers of two consecutive circles is \(34 + 34 = 68\).
- Since there are 8 circles, the total distance from the center of the first circle to the center of the last circle would be \(68 \times (8 - 1) = 68 \times 7 = 476\). However, we need to consider that the first circle is tangent to \(AB\) and the last circle is tangent to \(BC\). This distance of \(476\) actually represents the length of the tangent chain along the line connecting the points of tangency on \(AB\) and \(BC\).

**Relate to Triangle \(ABC\):**

- If we look at the centers of these circles, they form a line parallel to \(BC\) and \(AB\), and the distance between this line and either \(AB\) or \(BC\) is equal to the radius of the circles (34 for the larger circles, 1 for the smaller circles).

**Calculate the Lengths:**

- For the larger circles (radius 34):

- The total distance connecting the points of tangency on \(AB\) and \(BC\) is effectively the distance between the parallel line formed by the centers of these circles and the line \(BC\) or \(AB\), which is \(8 \times 34 = 272\) (since there are 8 circles, and each has a radius of 34, the total vertical distance covered would be the diameter of 7 circles if we think about it in terms of centers plus radius). However, since they are tangent to each other sequentially, the distance between the lines \(AB\) and \(BC\) (parallel to the line connecting the centers) should be \(34 + 34 \times (8 - 1) = 34 + 34 \times 7 = 34 \times 8 = 272\).

- For the smaller circles (radius 1):

- Similarly, if we can fit 2024 circles of radius \(1\), the total vertical distance from the line of tangency on \(AB\) to the line of tangency on \(BC\) would be \(2024 \times 1 = 2024\).

**Inradius Relationship:**

- The inradius \(r\) of triangle \(ABC\) is the radius of the inscribed circle. To find this, we need to understand how the heights and lengths of the triangle are affected by these tangent circles.

- Let’s denote the length of \(AB = c\), \(BC = a\), and \(CA = b\). The height from \(A\) to \(BC\) can be considered by looking at how the circles stack up vertically.

- For the larger circles:

- If two circles are tangent to \(AB\) and \(BC\), the length along \(AB\) and \(BC\) would create a configuration where if you think about the points of tangency, the distance covered by the centers vertically for 8 circles should match up with the triangle height minus twice the radius (since the circles are also on the boundary of the triangle).

\end{tabular}
\end{table*}

\begin{table}[!t]
    \centering
    \fontsize{9pt}{11pt}\selectfont
    \begin{tabular}{p{0.98\linewidth}}
    \midrule
        \rowcolor{gray!20}
        \textbf{Model Output by \algo-7B (Part-II)}\\
    \midrule
    
\textit{......Some Response Omitted......}

Given this, distance between sides should follow proportionally:
Distance apart: smaller = \(\frac{2024 \times 1}{8 \times 34} \)
Total height \(h\) of the triangle should include this detailed intermediary:

Let’s re-evaluate and verify it step-by-step via coding:
\begin{lstlisting}[style=python]
    import sympy as sp
# Define the radius of larger and smaller circles
R_large = 34
R_small = 1
num_large_circles = 8
num_small_circles = 2024
# Total vertical distance covered by the larger circles' centers from base to the point of tangency
distance_large_centers = num_large_circles * (2 * R_large)
# How many points can fit in this same height but smaller the vertical line can stack:
distance_small_centers = num_small_circles * (2 * R_small)
# Height from tangency to the actual top parallel line keeping them tangent:
height_large = distance_large_centers # actually includes the tangency start:
height_small = distance_small_centers
# Deriving the total height by proportionality:
R_ratio = R_small / R_large
num_ratio = num_small_circles / num_large_circles
HR_length = (num_large_circles * 2 * R_large) # covering space.
H_effective = HR_length
# This total covers height only parallelly: from base and top distance:
h_total = HR_length
inradius = height_large / (num_ratio + 1)
inradius = sp.Rational(h_total / (num_ratio + 1))
print(f"calculated inradius: {inradius}")
\end{lstlisting}
\red{Code execution result: calculated inradius: 4822752356869193/2251799813685248}\\
\midrule

This calculation seems overly complex; let’s simplify using simpler, geometric visualization and relationships:

Simplified Geometric Relationship:

**Large Circle Tangent Arrangement:**

- We have 8 circles of radius \(34\), and 2 of these circles are tangent to \(AB\) and \(BC\). The effective vertical distance taken by these circles, considering their arrangement repeating pattern of centers, is given by:
\[
8 \cdot (2 \cdot 34) = 8 \cdot 68
\]

**Small Circle Arrangement:**

- For smaller circles of radius \(1\), such that \(2024\) circles fit similarly:
- The vertical distance arrangement for \(2024\) circles would be:
\[
2024 \cdot 2 \cdot 1 = 2024 \cdot 2 = 4048
\]

\textit{......Some Response Omitted......}

We repeat arranging parallel:
Consider:

each pair arrangement, verifying circles top-height Even scan:
If each arrangement euclidean , height method involving polygons: solved for ratio translating through unique noting:
Combining parameters:
This par; valid:\\
\midrule

\textbackslash(\textbackslash frac\{m\}\{n\}\textbackslash] sum ratio sorts in idlegate ration 1\} . However, i' sequential seek refining this calculation more consistently: iterating setup considered overall expression precise:
Now we solve ensuring configurations understandable suggest circle distributions proportionally:
If):

- \textbackslash( \textbackslash( each step repeated configurations: compute \textbackslash(largest
Thus:
We need to find side cognize area further on: triangle holds hold altitude relation contractive step:
\end{tabular}
\end{table}

We present representative failure cases that contain void turns, i.e., turns that produce neither a complete, executable code block nor a boxed final answer. These examples serve a diagnostic role: they illustrate how OOD tool feedback and compounding errors precipitate collapsed generations and gradient spikes during Zero RL. Tab.~\ref{tab:append_incomplete} shows a typical trajectory in which a void turn disrupts subsequent decoding and leads to corrupted outputs, motivating our trajectory filtering rule.

\subsection{Response with Emergent Reasoning Behaviors}
\label{sec:append_reasoning_behavior}

\begin{table*}[!t]
    \centering
    \caption{An example from \algo-7B on AIME24 dataset. It demonstrates the behavior of progressive reasoning and code improvement. Responses generated in different turns are separated with a vertical line.}
    \label{tab:append_reasoning}
    \fontsize{9pt}{11pt}\selectfont
    \begin{tabular}{p{0.98\linewidth}}
    \midrule
        \rowcolor{gray!20}\textbf{Example \#1 from AIME24 Dataset} \\
    \midrule
        \textbf{Question:} \\
Let $ABCD$ be a tetrahedron such that $AB=CD=\sqrt{41}$, $AC=BD=\sqrt{80}$, and $BC=AD=\sqrt{89}$. There exists a point $I$ inside the tetrahedron such that the distances from $I$ to each of the faces of the tetrahedron are all equal. This distance can be written in the form $\frac{m\sqrt n}{p}$, where $m$, $n$, and $p$ are positive integers, $m$ and $p$ are relatively prime, and $n$ is not divisible by the square of any prime. Find $m+n+p$.\\
        \textbf{Labeled Answer:} \\
        104 \\
    \midrule
        \rowcolor{gray!20}
        \textbf{Model Output by \algo-7B (Part-I)}\\
    \midrule
To solve the given problem, we need to find the distance from point \( I \) (which is the incenter of the tetrahedron \( ABCD \)) to each of the faces of the tetrahedron, given that the distances are all equal. This distance is the radius \( r \) of the inscribed sphere of the tetrahedron.

First, we need to determine the volume \( V \) of the tetrahedron and the surface area \( S \). The formula for the radius \( r \) of the inscribed sphere is given by:

\[ r = \frac{3V}{S} \]

\#\#\# Step 1: Calculate the Volume \( V \) of the Tetrahedron

We can use the Cayley-Menger determinant to find the volume of the tetrahedron. The Cayley-Menger determinant for a tetrahedron with edge lengths \( a, b, c, d, e, f \) (where \( a = AB, b = AC, c = AD, d = BC, e = BD, f = CD \)) is given by:

\[
288 V^2 = \begin{vmatrix}
0 & 1 & 1 & 1 & 1 \\
1 & 0 & a^2 & b^2 & c^2 \\
1 & a^2 & 0 & d^2 & e^2 \\
1 & b^2 & d^2 & 0 & f^2 \\
1 & c^2 & e^2 & f^2 & 0
\end{vmatrix}
\]

Substituting \( a = \sqrt{41}, b = \sqrt{80}, c = \sqrt{89}, d = \sqrt{89}, e = \sqrt{80}, f = \sqrt{41} \):

\[
288 V^2 = \begin{vmatrix}
0 & 1 & 1 & 1 & 1 \\
1 & 0 & 41 & 80 & 89 \\
1 & 41 & 0 & 89 & 80 \\
1 & 80 & 89 & 0 & 41 \\
1 & 89 & 80 & 41 & 0
\end{vmatrix}
\]

We can calculate the determinant of the matrix:
        \begin{lstlisting}[style=python]
import sympy as sp

# Define the matrix
matrix = sp.Matrix([ [0, 1, 1, 1, 1], [1, 0, 41, 80, 89], [1, 41, 0, 89, 80], [1, 80, 89, 0, 41], [1, 89, 80, 41, 0]])

# Calculate the determinant
det = matrix.det()

# Volume squared
V_squared = det / 288

# Volume
V = sp.sqrt(V_squared)

print(f"Volume V: {V}")
\end{lstlisting}
\red{Code Execution Result: Volume V: 160/3}\\
\midrule
\end{tabular}
\end{table*}

\begin{table}[!t]
    \centering
    \fontsize{9pt}{11pt}\selectfont
    \begin{tabular}{p{0.98\linewidth}}
    \midrule
        \rowcolor{gray!20}
        \textbf{Model Output by \algo-7B (Part-II)}\\
    \midrule
\#\#\# Step 2: Calculate the Surface Area \( S \) of the Tetrahedron

The surface area \( S \) is the sum of the areas of the four triangular faces. We can find the area of each triangle using Heron's formula.

For triangle \( ABC \):

- \( AB = \sqrt{41} \)
- \( AC = \sqrt{80} \)
- \( BC = \sqrt{89} \)

The semi-perimeter \( s \) is:
\[ s = \frac{\sqrt{41} + \sqrt{80} + \sqrt{89}}{2} \]

Using Heron's formula:
\[ \text{Area} = \sqrt{s(s - AB)(s - AC)(s - BC)} \]

We need to find the areas of triangles \( ABD, ACD, \) and \( BCD \) similarly. Let's calculate these areas step by step using Python.
        \begin{lstlisting}[style=python]
import sympy as sp

# Define the edge lengths
AB = sp.sqrt(41)
AC = sp.sqrt(80)
AD = sp.sqrt(89)
BC = sp.sqrt(89)
BD = sp.sqrt(80)
CD = sp.sqrt(41)

# Function to compute the area of a triangle using Heron's formula
def triangle_area(a, b, c):
    s = (a + b + c) / 2
    area = sp.sqrt(s * (s - a) * (s - b) * (s - c))
    return area

# Calculate the areas of the triangular faces
area_ABC = triangle_area(AB, AC, BC)
area_ABD = triangle_area(AB, AD, BD)
area_ACD = triangle_area(AC, AD, CD)
area_BCD = triangle_area(BC, BD, CD)

# Total surface area
S = area_ABC + area_ABD + area_ACD + area_BCD

print(f"Area of triangle ABC: {area_ABC}")
print(f"Area of triangle ABD: {area_ABD}")
print(f"Area of triangle ACD: {area_ACD}")
print(f"Area of triangle BCD: {area_BCD}")
print(f"Total surface area S: {S}")
\end{lstlisting}
\red{Code Execution Result: Area of triangle ABC: sqrt(-sqrt(41)/2 + 2sqrt(5) + sqrt(89)/2)sqrt(sqrt(41)/2 + 2sqrt(5) + sqrt(89)/2)sqrt(-2sqrt(5) + sqrt(41)/2 + sqrt(89)/2)sqrt(-sqrt(89)/2 + sqrt(41)/2 + 2sqrt(5))
Area of triangle ABD: sqrt(-sqrt(41)/2 + 2sqrt(5) + sqrt(89)/2}

\red{...This content has been truncated to stay below 512 characters...}

\red{)/2 + sqrt(89)/2)sqrt(-sqrt(89)/2 + sqrt(41)/2 + 2sqrt(5))
Total surface area S: 4sqrt(-sqrt(41)/2 + 2sqrt(5) + sqrt(89)/2)sqrt(sqrt(41)/2 + 2sqrt(5) + sqrt(89)/2)sqrt(-2sqrt(5) + sqrt(41)/2 + sqrt(89)/2)sqrt(-sqrt(89)/2 + sqrt(41)/2 + 2sqrt(5))}\\
\midrule
It appears the individual areas calculated by Heron's formula are quite complicated and expressed in a form needing simplification to find the total surface area \( S \). Let’s simplify the expressions to make the calculations clearer and ensure we obtain a numerical value for the surface area \( S \). We can use simplify to get more manageable expressions and then compute the numerical value. Let’s also calculate \( r = \frac{3V}{S} \) step by step to simplify finding the equal distances. We know the volume \( V = \frac{160}{3} \).
\end{tabular}
\end{table}

\begin{table}[!t]
    \centering
    \fontsize{9pt}{11pt}\selectfont
    \begin{tabular}{p{0.98\linewidth}}
    \midrule
        \rowcolor{gray!20}
        \textbf{Model Output by \algo-7B (Part-III)}\\
    \midrule
We will now simplify our approach and focus on calculating these areas numerically to find \( S \). Afterward, we will compute \( r \) and ensure the result is in the required form \(\frac{m \sqrt{n}}{p}\). Let's simplify our strategy and calculate this accurately in Python.
        \begin{lstlisting}[style=python]
import sympy as sp
# Define symbolic variables
sqrt_41 = sp.sqrt(41)
sqrt_80 = sp.sqrt(80)
sqrt_89 = sp.sqrt(89)
# Function to compute area using Heron's formula
def area_heron(a, b, c):
    s = (a + b + c) / 2
    area = sp.sqrt(s * (s - a) * (s - b) * (s - c))
    return sp.simplify(area)
# Calculate areas of the triangles
area_ABC = area_heron(sqrt_41, sqrt_80, sqrt_89)
area_ABD = area_heron(sqrt_41, sqrt_89, sqrt_80)
area_ACD = area_heron(sqrt_80, sqrt_89, sqrt_41)
area_BCD = area_heron(sqrt_89, sqrt_80, sqrt_41)
# Total surface area
S = area_ABC + area_ABD + area_ACD + area_BCD
# Volume
V = sp.Rational(160, 3)
# Radius of the inscribed sphere
r = 3 * V / S
# Simplify r
r_simplified = sp.simplify(r)
print("Radius r:", r_simplified)
\end{lstlisting}
\red{Code Execution Result: Radius r: 20*sqrt(21)/63}\\

\midrule
The radius \( r \) of the inscribed sphere is \( \frac{20\sqrt{21}}{63} \).

This is in the form \( \frac{m \sqrt{n}}{p} \), where \( m = 20 \), \( n = 21 \), and \( p = 63 \). Here, \( m \) and \( p \) are relatively prime, and \( n \) is not divisible by the square of any prime.

\#\#\# Step 3: Finding \( m + n + p \)

Now, we need to calculate \( m + n + p \):

\[ m + n + p = 20 + 21 + 63 \]

Let’s calculate this:
        \begin{lstlisting}[style=python]
import sympy as sp

# Values
m = 20
n = 21
p = 63

# Sum
result = m + n + p
print(result)
\end{lstlisting}
\red{Code Execution Result: 104}\\

\midrule
The value of \( m + n + p \) is \( \boxed{104} \).

This completes the solution. The distance from point \( I \) to each of the faces of the tetrahedron is \(\frac{20 \sqrt{21}}{63}\), and the final answer is \( \boxed{104} \).
\end{tabular}
\end{table}

We provide qualitative rollouts that demonstrate the diverse multi-turn behaviors \algo elicits without instruction-level biases. Tab~\ref{tab:append_reasoning} illustrates progressive reasoning with code improvement. Taken together with the quantitative pattern analysis in the main text, these cases substantiate our claim that Zero RL with TIR encourages richer strategies than cold-start SFT.
\section{Experiments}

\subsection{Prompt for Multi-turn TIR Generation}

We include the exact prompt template used to generate multi-turn TIR trajectories in Tab.~\ref{tab:tir_prompt}. The design emphasizes: (1) selective use of Python wrapped in triple backticks as complete scripts (with imports); (2) explicit printing of intermediate quantities so that execution feedback can guide later turns; and (3) a standardized answer channel (\texttt{final\_answer(...)} or \textbackslash boxed\{...\}) that cleanly terminates trajectories when a solution is reached. These choices stabilize interaction with the interpreter, reduce format variance, and make it easy to detect valid tool calls versus void turns.
\subsection{Hyperparameters}
\label{sec:hyper}

We show the training hyperparameters of \algo in Tab.~\ref{tab:hyperparams-simpletir-2col}. Below we explain the rationale behind the hyperparameters. We cap the initial max response length at 16,384 tokens to accommodate complete code blocks and verbose execution traces without premature truncation. Initial max interaction turns = 5 bounds episode length and compute while still allowing the model to plan, execute, and verify within a single trajectory. We set rollout temperature = 1.0 to preserve diversity in candidate solutions and rely on selection/credit assignment rather than explicit entropy bonuses to drive exploration. Each update uses a sampling batch size of 1,280 responses with n = 16 rollouts per prompt, which yields broad coverage of tool-use strategies per input while keeping variance manageable.

We use standard PPO with clip ratio = 0.2 / 0.28 (low/high) to constrain policy updates; the slightly looser upper bound avoids over-penalizing advantageous moves identified by execution feedback. PPO epochs = 4 provide sufficient reuse of samples without overfitting to batch noise. The train batch size = 512 balances gradient estimate quality and memory use. A small actor learning rate = 1e-6 and gradient clipping = 1 (global-norm) prevent instability from long-context backprop and occasional high-magnitude advantages produced by sparse rewards. We treat each dialogue as an undiscounted episodic task with $\gamma=1.0$. We also set entropy coefficient = 0 and KL coefficient = 0. Exploration is instead induced by temperature and multi-rollout sampling.

\begin{table*}[t]
\centering
\caption{Key hyperparameters (left) and the prompt for generating multi-turn TIR responses (right).}
\label{tab:hyperparams_and_prompt}
\begin{subtable}[t]{0.42\textwidth}
\centering
\small
\caption{Key Hyperparameters}
\label{tab:hyperparams-simpletir-2col}
\begin{tabular}{@{}ll@{}}
\toprule
\textbf{Hyperparameter} & \textbf{Value} \\
\midrule
Initial max response length & \texttt{16384} \\
Rollout Temperature & \texttt{1} \\
Initial max interaction turns & \texttt{5} \\
Train batch size & \texttt{512} \\
Sampling batch size & \texttt{1280} \\
Rollouts per prompt ($n$) & \texttt{16} \\
PPO clip ratio (low / high) & \texttt{0.2 / 0.28} \\
Entropy coefficient & \texttt{0} \\
Discount factor $\gamma$ & \texttt{1.0} \\
GAE $\lambda$ & \texttt{1.0} \\
KL coefficient ($\beta$) & \texttt{0} \\
PPO epochs & \texttt{4} \\
Actor learning rate & \texttt{1e-6} \\
Gradient Clipping & \texttt{1} \\
\bottomrule
\end{tabular}
\end{subtable}
\hfill
\begin{subtable}[t]{0.55\textwidth}
\centering
\caption{The Prompt for Generating Multi-turn TIR Responses}
\label{tab:tir_prompt}
\fontsize{9pt}{11pt}\selectfont
\begin{tabular}{p{0.98\linewidth}}
\toprule
Solve the following problem step by step. You now have the ability to selectively write executable Python code to enhance your reasoning process. The Python code will be executed by an external sandbox, and the output (after ``Code execution result: '') is returned to aid your reasoning and help you arrive at the final answer. The Python code should be complete scripts, including necessary imports. 
    
\medskip
\textbf{Code Format:} \\
Each code snippet is wrapped between \`{}\`{}\`{}. You need to use \texttt{print()} to output intermediate results.

\medskip
\textbf{Answer Format:} \\
You can use the \texttt{final\_answer()} function in the code to return your final answer. For example, to answer the User Question: What is the result of the 5 + 3 + 1294.678?, you can write:
\begin{lstlisting}[style=python]
answer = 5 + 3 + 1294.678
final_answer(answer)
\end{lstlisting}

You can also use \textbackslash boxed to return your answer. The last part of your response should be: 
\textbackslash boxed\{``The final answer goes here.''\}

\medskip
\textbf{User Question:}\\
\bottomrule
\end{tabular}
\end{subtable}

\end{table*}

\subsection{Prompt for Extracting Reasoning Patterns}
\label{sec:prompt_reasoning}
\begin{table*}[h]
    \centering
    \caption{The prompt that instructs Claude-3.7-Sonnet to extract reasoning patterns from the TIR trajectories.}
    \label{tab:prompt_reasoning}
    \fontsize{9pt}{11pt}\selectfont
    \begin{tabular}{p{0.98\linewidth}}
    \toprule
I have a reasoning process of an LLM. The LLM can write code and get code execution result. According to the following reasoning process, please first answer the following questions:
\\

1. Is the code execution result or interpreter output equal to the final answer?

2. How many code blocks are there in the reasoning process?

3. If there are several code blocks, are the code execution results all the same?

\\
Format:

1. xxx

2. xxx

3. xxx

\\
Please then determine whether the following reasoning process contains following four reasoning patterns:

1. Include at least two code blocks, each solving unique sub-questions. **Important: in such case, the code execution result or interpreter output should not be equal to the final answer**

2. Use induction, from special case to general conclusions

3. Use code or text to do self-verification

4. Write another code block when the previous code has some bugs

\\
Format:

Reasoning Pattern 1: Yes/No

Reasoning Pattern 2: Yes/No

Reasoning Pattern 3: Yes/No

Reasoning Pattern 4: Yes/No

\\
Please do not output any other words.

\\
Reasoning process:\\
    \bottomrule
    \end{tabular}
\end{table*}

To analyze behaviors systematically, in Tab.~\ref{tab:prompt_reasoning} we provide the prompt used to label reasoning behaviors in trajectories . It first enforces minimal bookkeeping (e.g., number of code blocks, whether any execution equals the final answer), then queries four patterns: decomposed sub-solves (with non-final execution), induction, self-verification, and bug-fixing via additional code. The constrained output format improves label consistency and allows fair frequency comparisons across methods.
\appendix
\end{document}